\documentclass[a4paper]{article}

\usepackage{INTERSPEECH2020}

\usepackage{xcolor}
\usepackage{hyperref}

\newcommand{\insertcovost}{
    \begin{table*}[t]
        \begin{center}
            \caption{BLEU on four language pairs of CoVoST-V2: English-German (en-de), English-Catalan (en-ca), English-Arabic (en-ar) and English-Turkish (en-tr).
            The results show that self-supervised pre-training on Librispeech (wav2vec-2.0) followed by self-training on the same data can improve performance. 
            Using a language model during decoding improves performance further.}
            \small
            \begin{tabular}[b]{l|l|cccc|c}
            \toprule
            {\bf Row} & {\bf Model} & {\bf en-de} & {\bf en-ca} & {\bf en-ar} & {\bf en-tr} & {\bf Avg}\\
                \midrule
                \midrule
                \multicolumn{7}{l}{\it Baselines from CoVoST 2} \\
                \midrule
                1 & Wang et al. \cite{wang2020covost2} (w/o pre-ASR) & 13.6 & 20.2 & 8.7 & 8.9 & 12.9 \\
                2 & Wang et al. \cite{wang2020covost2} (w/ pre-ASR) & 16.3 & 21.8 & 12.1 & 10.0 & 15.1 \\
                \midrule
                \midrule
                \multicolumn{7 }{l}{\it Previous state-of-the-art results} \\
                \midrule
                3 & Prev E2E SOTA \cite{iranzo2020europarl,wang2020covost2,li2021multilingual} & 18.4 & 23.6 & 13.9 & 11.7 & 16.9 \\
                4 & Cascade SOTA \cite{li2021multilingual} & 19.4 & 25.0 & 14.3 & 11.7 & 17.6 \\
                5 & Li et al. \cite{li2021multilingual} (joint training) & 25.8 & 30.9 & 18.0 & 17.0 & 22.9 \\
                6 & Li et al. \cite{li2021multilingual} (+ extra MT data) & 26.6 & 30.4 & 18.6 & 16.3 & 23.0 \\
                \midrule
                \midrule
                \multicolumn{7}{l}{\it Our results} \\
                \midrule
                8 & wav2vec-2.0 & 23.8 & 32.4 & 17.4 & 15.4 & 22.3 \\
                9 & wav2vec-2.0 + decoding w/ LM & 24.9 & 34.0 & 18.0 & 16.7 & 23.4 \\
                10 & wav2vec-2.0 + self-training (LV-60k) & 26.5 & 34.1 & 20.2 & 17.5 & 24.6 \\
                11 & wav2vec-2.0 + self-training (LV-60k) + decoding w/ LM & \bf 27.2 & \bf 35.6 & \bf 20.8 & \bf 18.9 & \bf 25.6 \\
                \bottomrule
            \end{tabular}
            \label{tab:main_results}
        \end{center}
       \vspace{-0.3cm}
    \end{table*}
}

\newcommand{\insertcovostdata}{
    \begin{table}[t]
        \begin{center}
            \caption{\textbf{CoVoST 2 dataset.} We use four language pairs of the CoVoST 2 dataset for our experiments, all from English due to the easy access of open-source English unnanotated data in that source language. 
            \label{tab:covost_data}}
            \resizebox{1\linewidth}{!}{
            \begin{tabular}[b]{l|cccc}
            \toprule
            Language pair & {\bf en-de} & {\bf en-ca} & {\bf en-ar} & {\bf en-tr} \\
                \midrule
                \midrule
            Speech hours & 430h & 430h & 430h & 430h \\
            Target utterances & 288K & 288K & 288K & 288K \\
            Target words & 2.8M & 3.1M & 2.4M & 2.2M \\
                \bottomrule
            \end{tabular}
            }
        \end{center}
       \vspace{-0.3cm}
    \end{table}
}

\newcommand{\insertdataablation}{
    \begin{table*}[t]
        \begin{center}
            \caption{Effect of the amount of unlabeled data on pretraining and self-training (cf.~\autoref{tab:main_results}).}
            \begin{tabular}[b]{l|l|cccc|c}
            \toprule
            {\bf Row} & {\bf Model} & {\bf en-de} & {\bf en-ca} & {\bf en-ar} & {\bf en-tr} & {\bf Avg}\\
                \midrule
                \midrule
                \multicolumn{7}{l}{\it Increasing the amount of pretraining data} \\
                \midrule
                \midrule
                1 & wav2vec-2.0 (LS-960) & 20.5 & 27.3 & 15.4 & 14.0 & 19.3 \\
                2 & wav2vec-2.0 (LV-60k) & 23.8 & 32.4 & 17.4 & 15.4 & 22.3 \\
                \midrule
                \midrule
                \multicolumn{7}{l}{\it Increasing the amount of self-training data} \\
                \midrule
                \midrule
                3 & wav2vec-2.0 + Self-training (LS-960)  & 25.4 & 32.9 & 19.1 & 16.4 & 23.5 \\
                4 & wav2vec-2.0 + Self-training (LV-6k)  & 25.9 & 33.2 & 19.9 & 17.5 & 24.1 \\
                5 & wav2vec-2.0 + Self-training (LV-60k)  & 26.5 & 34.1 & 20.2 & 17.5 & 24.6 \\
                \bottomrule
            \end{tabular}
            \label{tab:ablation_size}
        \end{center}
       \vspace{-0.3cm}
    \end{table*}
}

\title{Large-Scale Self- and Semi-Supervised Learning for Speech Translation}
\name{$^1$}

\name{Changhan Wang$^*$, Anne Wu$^*$, Juan Pino$^*$, Alexei Baevski, Michael Auli, Alexis Conneau}
\address{
  Facebook AI}
\email{\{changhan,juancarabina,abaevski,michaelauli,aconneau\}@fb.com,annewu018@gmail.com}

\begin{document}

\maketitle

\begin{abstract}
In this paper, we improve speech translation (ST) through effectively leveraging large quantities of unlabeled speech and text data in different and complementary ways. We explore both pretraining and self-training by using the large Libri-Light speech audio corpus and language modeling with CommonCrawl. Our experiments improve over the previous state of the art by 2.6 BLEU on average on all four considered CoVoST 2 language pairs via a simple recipe of combining wav2vec 2.0 pretraining, a single iteration of self-training and decoding with a language model. Different to existing work, our approach does not leverage any other supervision than ST data. Code and models will be publicly released.

\end{abstract}
\noindent\textbf{Index Terms}: speech translation, unsupervised pretraining, self-training, semi-supervised learning

\renewcommand{\thefootnote}{$^*$}
\footnotetext[1]{Equal contribution.}
\renewcommand\thefootnote{\arabic{footnote}}

\section{Introduction}

Similar to many machine learning tasks, the amount of labeled data for speech-to-text applications such as automatic speech recognition (ASR) and speech translation (ST) is limited. 
For speech and language tasks, this problem is exacerbated by the fact that people speak many languages around the world and obtaining large quantities of labeled speech translation data for every language is simply not practical.
On the other hand, unlabeled speech audio or text data is much more plentiful and various techniques to utilize it have been explored.

Semi-supervised learning techniques for ASR such as unsupervised speech pretraining~\cite{oord2018representation,Schneider2019,baevski2020wav,conneau2020unsupervised} and self-training or (iterative) pseudo-labeling~\cite{DBLP:journals/corr/abs-1911-08460,kahn2020self,xu2020iterative,likhomanenko2020slimipl} have substantially improved performance on the traditional LibriSpeech benchmark, and led to systems that can learn with little supervision.
In addition, it was recently shown that self-training and self-supervised pretraining can be effectively combined~\cite{xu2020selftraining,zhang2020pushing} for speech recognition to achieve 4.8 WER on LibriSpeech test-other with only 10 minutes of annotated speech data.

Similarly for speech translation, there has been work on combating data scarcity, namely multitask learning~\cite{weiss2017sequence,berard2018endtoend,tang2020general}, pretraining on ASR data~\cite{berard2018endtoend,bansal-etal-2019-pre,stoian2020analyzing,wang2020bridging}, data augmentation~\cite{jia2019leveraging,pino2019harnessing,salesky-etal-2019-exploring,mccarthy2020skinaugment}, self-supervised pretraining~\cite{Wu2020,Nguyen2020}, self-training~\cite{Pino2020} or multilingual speech translation~\cite{gangi2019onetomany,inaguma2019multilingual,wang-EtAl:2020:LREC1,wang2020covost2,li2021multilingual}. 
However, multitask learning, pretraining, data augmentation and multilingual ST rely on additional supervision provided by labeled ASR data, machine translation (MT) data or ST data while self-training and self-supervised pretraining exploit unlabeled source speech data. 
Unlike in ASR, the complementarity of self-supervised learning and self-training (and other semi-supervised techniques) has not been studied for the ST task.
%\juan{what's missing here is some citations of ST work that uses monolingual target data}

% For ASR, it has been recently shown that pretraining and self-training are complementary even when both methods use the same unlabeled data~\cite{xu2020selftraining}.
% While both self-supervised pretraining and self-training have been explored individually for speech translation, there has be no work which studied their combined effects. One of the contributions of this paper is to explore whether they are complementary or not in the context of speech translation

On the other hand, past work in ST relied heavily on labeled data in the form of speech-to-text translation data, ASR transcriptions~\cite{livescu2019pre} or MT sentence pairs to improve performance~\cite{li2021multilingual}. 
% \michael{This point overlaps with the comment about multitask learning etc. using additional supervision. Can we consolidate this somehow?}
In this work, we follow the trend of leveraging purely unlabeled data to improve performance and show that this obtains strong performance while leveraging only supervision from ST data. Using wav2vec 2.0 pretraining, self-training and language model decoding, we show that we can outperform previous work while leveraging much less supervision.

Our contributions are as follows: we present a comprehensive study of the impact of existing semi-supervised learning techniques on speech translation and show that they greatly reduce the need for additional supervision in the form of labeled ASR or translation parallel data. We show that our simple approach obtains state-of-the-art results on all four language pairs we evaluate on: English to German, English to Catalan, English to Arabic and English to Turkish. We also conduct an ablation study on the impact of the quantity of unlabeled data for self-training and self-supervised pre-training in the context of ST.

In what follows, we describe the semi-supervised learning techniques and the system we use for ST. Then we present our results on the CoVoST 2 benchmark~\cite{wang2020covost2} on four language pairs and compare our work to the literature.
%We also conduct an analysis on learning with a smaller amount of supervision.
% Finally, we conclude with future directions for semi-supervised learning in ST.

\section{Learning from Unlabeled Data}
In this section, we describe the techniques to leverage unlabeled speech or text data which we use in this study.

\subsection{Unsupervised Pretraining}

Unsupervised pretraining has been very effective in multiple fields of machine learning, including natural language processing~\cite{devlin2018bert}, computer vision~\cite{he2019momentum,chen2020simple} and speech recognition~\cite{baevski2020wav}. 
In this work, we demonstrate the impact of unsupervised pretraining for speech translation (ST) by leveraging a wav2vec 2.0 model pretrained on Libri-Light\footnote{\scriptsize\url{github.com/pytorch/fairseq/blob/master/examples/wav2vec}}, a unlabeled dataset comprising 53K hours of English read audio books.  
The model is trained by predicting the latent speech representations of masked time-steps using a loss similar to SimCLR~\cite{chen2020simple}.
The latent speech representations are quantized for the prediction task and there is a fixed number of latents stored in a codebook.

For ST, we construct a sequence-to-sequence model with attention~\cite{sutskever2014sequence,chorowski2015attention} by adding a randomly initialized decoder model on top of a wav2vec 2.0 encoder. 
After pretraining, all parameters are fine-tuned on the CoVoST 2 ST data~\cite{wang2020covost2}. The decoder is also a Transformer model but smaller, with 7 layers and model dimension 256, which we do not pretrain\footnote{Pretraining with a masked language model did not improve performance in our setting}.

\subsection{Self-Training}
Self-training is a semi-supervised learning method that first trains a \textit{teacher model} on labeled data.
The teacher model is then used to synthetically annotate unlabeled data in order to train a new \textit{student model} on the combination of labeled and pseudo-labeled data~\cite{park2020improved,Pino2020,kahn2020st}.
% Iterative self-training such as the NoisyStudent approach iterates that process several times until convergence~\cite{xu2020iterative,park2020improved}.
We further fine-tune the student model on labeled data to alleviate the domain mismatch between labeled and unlabeled data.
Recent work showed that unsupervised pretraining and self-training can be complementary for natural language understanding~\cite{du2020self} and speech recognition~\cite{xu2020selftraining}. In this work, we adopt this setting by considering self-training on top of a wav2vec 2.0 pretrained model and show the complementarity of these two learning approaches for speech translation.

\subsection{Decoding with Language Model}
With unsupervised pretraining and self-training, we leverage additional unlabeled speech data to improve the performance of a ST system. We also make use of monolingual text data in the target language to further improve translation quality. Specifically, we train a language model (LM) on part of the CommonCrawl data\footnote{\url{http://data.statmt.org/cc-100/}} that is in similar domains as CoVoST 2 (\autoref{sec:datasets}). And then we combine ST model and LM scores at every time step in beam search decoding (shallow fusion)~\cite{gulcehre2015using}. %\juan{maybe shallow fusion by Gulcehre et al 2015 and Simple Fusion: Return of the Language Model by Stahlberg et al. ?}

% Unlike machine translation, where back-translation directly aims at improving the conditional probably of the target sentence given the source sentence, language model decoding only improves modeling of the target sentence. Back-translation is much more difficult to implement in speech translation given the asymmetry between text and speech: building a backward text-to-speech translation model is a much harder task. \juan{not clear why hard: let's explain that either you build a direct source text to translated audio system, which has not been done before AFAIK, or you use a cascaded backtranslation: MT backtranslation followed by TTS; the difficulty is that the TTS has to be multispeaker, ow it doesn't work; so yes, it's difficult!} In our work, we combine self-training and a target language-model, that is we improve the generation of the teacher model using language modeling, and train the student model on the output.  This combines the use unannotated data from the source and target sides to improve the conditional probability of the target sentence given the source speech audio.

% ALEXIS: CONTINUE FROM HERE

\section{Datasets and Training Details}
In this section, we describe the datasets we use for speech translation (ST), the setups for unsupervised pretraining, self-training as well as language modeling. Then we give details on how we train our models.

\subsection{Datasets}
\label{sec:datasets}

\insertcovostdata

\textbf{CoVoST 2: speech translation data.} \ CoVoST 2\footnote{\url{https://github.com/facebookresearch/covost}} is a large-scale multilingual speech translation corpus covering translations from 21 languages into English and from English into 15 languages. This represents the largest open dataset available to date from total volume and language coverage perspective. Specifically, we cover four language directions from English to German (de), Catalan (ca), Arabic (ar) and Turkish (tr), which contain 430 hours of annotated data each. For simplicity, we choose to focus on speech translation from English because of easy public access to unlabeled data in that language. 
% We found that using this model led to significantly better performance (+XX BLEU on average) than using a wav2vec 2.0 model directly trained on the ~400h speech data of the CoVoST dataset.
% AC: Add a link to the data?
\\

\noindent\textbf{Libri-Light: pretraining and self-training data.} \ We use the wav2vec 2.0 model\footnote{\scriptsize\url{dl.fbaipublicfiles.com/fairseq/wav2vec/wav2vec_vox_new.pt}} pretrained on the Libri-Light data~\cite{librilight}, a dataset consisting of more than 60k hours of unlabeled read speech data\footnote{\url{https://github.com/facebookresearch/libri-light}}. The dataset is derived from open-source audio books from the LibriVox project and is the largest freely-available corpus of speech. Previous work in unsupervised pretraining showed that using Libri-Light over LibriSpeech led to better performance~\cite{baevski2020wav}. We use the 6000-hour subset of Libri-light (LV-6k) as well as the full 60k hours dataset (LV-60k) for self-training. Specifically, we train models on CoVoST 2 whose domain is read speech, and synthetically label unannotated data from LV-60k - which is also read speech. We then fine-tune a wav2vec 2.0 pretrained model - the student model -  on both the synthetically annotated data and the ground-truth CoVoST 2 training data. Self-training is a good alternative to back-translation in the case of speech translation, although the target data generated by the teacher model is not real data. To remedy this issue, we leverage ground-truth target data through language model decoding. Note that similar to NoisyStudent, we also inject noise in our input during self-training through the same masking strategy as wav2vec 2.0~\cite{park2019specaugment,baevski2020wav}.
\\

\noindent\textbf{LibriSpeech: pretraining and self-training ablation data.} \ In order to study the effect of the amount of unlabeled data on both pretraining and self-training, we use the 960 hours of LibriSpeech as a smaller-scale alternative to Libri-light and LV-60k for wav2vec 2.0 pretraining\footnote{\scriptsize\url{dl.fbaipublicfiles.com/fairseq/wav2vec/wav2vec_small.pt}} and for self-training.
\\

\insertcovost

\noindent\textbf{CommonCrawl: language model data.} \ In order to obtain language model data in the right domain, we leverage CommonCrawl (CC) data from the CC100 corpus~\cite{conneau-etal-2020-unsupervised,wenzek-etal-2020-ccnet} for the four target languages studied (German, Catalan, Arabic, Turkish). First, we train 4-gram language models with the KenLM toolkit~\cite{heafield-etal-2013-scalable} on the training set of the CoVoST 2 data and on the CC data and filter CC by averaging the LM scores~\cite{moore2010intelligent}. We then keep only one tenth of the original data, 
% \juan{how many words/sentences}\changhan{The files were in Anne's folder which have been gone now?} 
and use transformer-based language models trained on both the CoVoST 2 training set and the additional CC data for language model rescoring.

\subsection{Training Details}
% Describe how we do speech translation with a sequence to sequence model (add the details of wav2vec 2.0 there, point to the code, like 
\noindent\textbf{Unsupervised pretraining.}
For wav2vec 2.0 pretraining, we use the \textit{Large} model, which comprises 24 self-attention blocks with model dimension 1024, inner dimension 4096 and 16 attention heads, resulting in a total of about 300M parameters. The feature encoder contains seven blocks and the temporal convolutions in each block have 512 channels with strides (5,2,2,2,2,2,2) and kernel widths (10,3,3,3,3,2,2), resulting in a receptive field of about 25ms and a stride of about 20ms. 

\noindent\textbf{Speech translation.}
We use a sequence-to-sequence model where the encoder is a wav2vec 2.0 model with several layers of convolutions followed by a Transformer network with 24 layers. 
The decoder is also a Transformer network with 7 layers, an embedding size of 256, 4 attention heads and FFN dimension of 2048.
% \michael{Above, in section "Unsupervised pretraining" it says the decoder has 6 layers and model dim 512. Which one is the correct setting?}
A 10K BPE vocabulary is built on the CoVoST 2 target text for each target language. We train our model with Adam, a learning rate of 5e-5, label smoothing with probability 0.1, an effective batch size of 6.4M tokens, layer drop 0.05, a masking strategy similar to wav2vec 2.0 with mask length 5 and mask probability 0.15. During the fine-tuning phase, the wav2vec 2.0 encoder is frozen for 10K updates. Models are trained for 250K updates and the best checkpoint is selected based on the BLEU score on the validation set. For self-training, we use a learning rate of 3e-5 and the same setting for the remaining hyperparameters. The pseudo-labels are generated with a beam size of 4.

\noindent\textbf{Language models.}
For the language models, we use the Google Billion Word Transformer architecture of~\cite{baevski2018adaptive} with 12 decoder layers and an embedding dimension of 512. We train the model with Adam, and an inverse sqrt scheduler. When using LM scores in decoding, we scale the scores by 0.1 and use a length penalty of 0.7.

% \juan{TODO(Changhan): add details about self-training setup}
% \changhan{I updated the paragraph above and the description in sec 2.2.}

\insertdataablation

\section{Experiments and Results}
In this section, we describe our experimental results obtained in Table~\ref{tab:main_results} and \ref{tab:ablation_size} where we combine self-supervised and semi-supervised learning techniques.
% \\
\subsection{Improvements from Unsupervised Pretraining}
We observe strong gains using wav2vec 2.0 models compared to previous baselines which were using similar speech translation architectures without pretraining.
% In particular, going from a model trained on LibriSpeech (960 hours) to Libri-Light (53K hours) leads to an average performance gain of 0.6 BLEU. \juan{this should be moved to the ablation section}
The Libri-light pretrained wav2vec 2.0 model (row 8) achieves 22.3 BLEU on average, which is on average 9.4 BLEU points better than the baseline of~\cite{wang2020covost2} (row 1) and 7.2 BLEU higher than their model which leveraged ASR pretraining as additional supervision (row 2). These results demonstrate that the features learned by the wav2vec 2.0 model are very useful beyond speech recognition, and applicable to other speech tasks such as speech translation. 
The results with only wav2vec 2.0 pretraining (row 8) come very close to the most recent state-of-the-art results (row 6) on CoVoST 2~\cite{li2021multilingual} which obtained 23 BLEU and even leads to a new state of the art on English-Catalan (row 8), without using any other supervision than the speech translation data.
%In this work, we only investigate how far we can push speech translation with only speech translation data and unannotated data. Our baseline thus leads to strong BLEU without any other type of supervision.
% \\

\subsection{Improvements from Self-Training}
As previously shown for computer vision~\cite{zoph2020rethinking}, natural language understanding~\cite{du2020self} and speech recognition~\cite{xu2020selftraining}, supervised or unsupervised pretraining can be complementary to self-training. 
Combining self-supervised learning with semi-supervised learning for speech translation in this work, we first use the previously described Libri-Light wav2vec 2.0 models fine-tuned on the CoVoST 2 speech translation data as teacher models, and synthetically annotate the 60k-hour Libri-light dataset (LV-60k).
%and CoVoST training data. 
We then leverage self-training by finetuning a wav2vec 2.0 pretrained student model on both LV-60k synthetic data and the CoVoST data~\cite{he2019revisiting,Pino2020}. Note that we upsample the ground-truth CoVoST data such that it has the same importance during finetuning than the synthetically annotated LV-60k. 
%In the combined We found that fine-tuning this model again on the ground-truth CoVoST data after that step leads to an additional average gain of 1.5 BLEU, confirming previous findings~\cite{he2019revisiting,Pino2020}.
After following this procedure, we obtain 24.6 average BLEU score (row 10) which is 2.3 BLEU better than using wav2vec 2.0 pretraining only, demonstrating the complementarity of pretraining and self-training for speech translation. We also observe similar level of improvements across language pairs which shows the consistency of this approach. With this method, we reach a new state of the art on the CoVoST 2 benchmark for Catalan, Arabic and Turkish (row 10).
% We show in Table~\ref{tab:covost} that self-training improves unsupervised pretraining for speech translation by 1.6 BLEU on average. This result is in line with previous work that showed the complementarity of pretraining and self-training~\cite{}.
% \\

\subsection{Improvements from Decoding with Language Model}
Self-training and unsupervised pretraining both leverage additional unannotated speech data to improve performance. But self-training generates noisy output on which the student model is fine-tuned which may lead the model to learn incorrect patterns.
To inject more prior knowledge about the target language structure, a natural solution is to use unannotated text in the target domain. In this work, we leverage language modeling as one way to do that, and use it to improve generation through decoding.
This improves the wav2vec 2.0 baseline by 1.1 BLEU on average across all language pairs (row 9 vs.\ row 8). 
It also improves the stronger setting of wav2vec 2.0 + self-training by 1 BLEU (row 11 vs.\ row 10). Combining wav2vec 2.0 pretraining, self-training and language model decoding in row 11, we reach a new state of the art on the CoVoST 2 benchmark, with an average BLEU score of 25.6 over the four language pairs.

% they do not leverage ground-truth target unannotated data to improve performance. As a way to leverage unannotated data on the target side, we apply language model decoding and observe a 2 BLEU improvement over our wav2vec baseline. Table ~\ref{tab:examples} illustrates the improved fluency of the generated sentences after language model decoding. Note that we tried pretraining the Transformer decoder with a language model but it did not improve performance.
% \\

% \subsection{Complementarity of Unsupervised Pretraining and Self-Training}

% In the line "self-train(wav2vec 2.0) + LM-dec", we apply self-training to our wav2vec 2.0 baseline, and apply LM-decoding afterwards.
% We show that it leads to complementary gains, reaching 19.7 BLEU on average, which is a 3.6 BLEU improvement over our strong baseline. Self-training combined with language model decoding can be seen as an alternative to back-translation in speech translation where backward text-to-speech models are difficult to train.
% \\

\subsection{Comparison with Previous Work}

The combination of pretraining, self-training and LM decoding outperforms the prior state of the art~\cite{li2021multilingual} in all language directions and by 2.6 BLEU on average. 
The prior state of the art uses both pretraining with wav2vec 2.0 and mBART, as well as a minimalistic LNA
(LayerNorm and Attention) finetuning. We note that mBART was fine-tuned using additional labeled machine translation data and therefore our approach relies on less labeled data. Moreover, this work is conducted on a bilingual setting without additional supervision coming from multiple language pairs. Finally, the prior state-of-the-art model uses an adapter between the encoder and the decoder that further downsamples the input by a factor of 8. In contrast, our model architecture was simplified by removing the adapter module, as well as the LNA finetuning.

\subsection{Data Ablation for Unsupervised Pretraining and Self-Training}

Prior work has shown that increasing the amount of unlabeled data for pretraining and self-training can improve ASR performance~\cite{baevski2020wav,xu2020selftraining}.
To understand whether the same holds true for ST, we compare pretraining a wav2vec 2.0 model on the 960 hours unlabeled speech audio of the smaller LibriSpeech corpus (LS-960) instead of the 60k hours of the LibriVox corpus (LV-60k) used so far. \autoref{tab:ablation_size} shows that pretraining the speech encoder on more data leads to a large improvement of 3 BLEU on average across the four language pairs (row 2 vs.\ row 1). Note that pretraining wav2vec 2.0 on LibriSpeech still provides an average improvement of 4.2 BLEU over the supervised baseline with ASR pretraining (row 1 vs.\ row 2 from \autoref{tab:main_results}).

Next, we examine increasing the amount of unlabeled data for self-training. We compare pseudo-labeling LS-960 with pseudo-labeling LV-6k and LV-60k. From \autoref{tab:ablation_size} we can see that the increase of self-training data from LS-960 to LV-6k brings an additional gain of 0.6 BLEU on average (row 4 vs.\ row 3). Scaling self-training even further to LV-60k leads to an additional gain of 0.5 BLEU on average (row 5 vs.\ row 4).  For both pretraining and self-training, we found that gains using more unlabeled data were similar when using a language model. 

% \juan{as discussed we want to say that the ablation holds true in the presence of LM decoding}

\section{Conclusion}

We pushed the limits of self-supervised and semi-supervised learning for speech translation by leveraging pretraining with wav2vec 2.0 and self-training. 
These techniques can outperform the previous state of the art by an average of 1.3 BLEU across four language directions without using any type of supervision other than the CoVoST 2 data. We also demonstrated the complementarity of unsupervised pretraining, self-training and language model decoding, outperforming previous approaches by 2.6 BLEU. Our work provides stronger and simpler baselines for speech translation and demonstrates the effectiveness of wav2vec 2.0 unsupervised pretraining for speech translation.

% \newpage
{
% \tiny
\bibliographystyle{IEEEtran}
\bibliography{interspeech}
}

% \begin{thebibliography}{9}
% \bibitem[1]{Davis80-COP}
%   S.\ B.\ Davis and P.\ Mermelstein,
%   ``Comparison of parametric representation for monosyllabic word recognition in continuously spoken sentences,''
%   \textit{IEEE Transactions on Acoustics, Speech and Signal Processing}, vol.~28, no.~4, pp.~357--366, 1980.
% \bibitem[2]{Rabiner89-ATO}
%   L.\ R.\ Rabiner,
%   ``A tutorial on hidden Markov models and selected applications in speech recognition,''
%   \textit{Proceedings of the IEEE}, vol.~77, no.~2, pp.~257-286, 1989.
% \bibitem[3]{Hastie09-TEO}
%   T.\ Hastie, R.\ Tibshirani, and J.\ Friedman,
%   \textit{The Elements of Statistical Learning -- Data Mining, Inference, and Prediction}.
%   New York: Springer, 2009.
% \bibitem[4]{YourName17-XXX}
%   F.\ Lastname1, F.\ Lastname2, and F.\ Lastname3,
%   ``Title of your INTERSPEECH 2020 publication,''
%   in \textit{Interspeech 2020 -- 20\textsuperscript{th} Annual Conference of the International Speech Communication Association, September 15-19, Graz, Austria, Proceedings, Proceedings}, 2020, pp.~100--104.
% \end{thebibliography}

\end{document}